\documentclass[runningheads]{llncs}
\usepackage[T1]{fontenc}
\usepackage{subfig}
\usepackage{graphicx}
\begin{document}
\title{Pretrained Visual Representations in Reinforcement Learning}
%
%
\author{Emlyn Williams \and
Athanasios Polydoros}
\authorrunning{Williams and Polydoros}
%
\institute{University of Lincoln, Lincoln, UK \\
\email{\{emwilliams, apolydoros\}@lincoln.ac.uk}
}
\maketitle              
\begin{abstract}
 Visual reinforcement learning (RL) has made significant pro-\\gress in recent years, but the choice of visual feature extractor remains a crucial design decision. This paper compares the performance of RL algorithms that train a convolutional neural network (CNN) from scratch with those that utilize pre-trained visual representations (PVRs). We evaluate the Dormant Ratio Minimization (DRM) algorithm, a state-of-the-art visual RL method, against three PVRs: ResNet18, DINOv2, and Visual Cortex (VC). We use the Metaworld Push-v2 and Drawer-Open-v2 tasks for our comparison. Our results show that the choice of training from scratch compared to using PVRs for maximising performance is task-dependent, but PVRs offer advantages in terms of reduced replay buffer size and faster training times. We also identify a strong correlation between the dormant ratio and model performance, highlighting the importance of exploration in visual RL. Our study provides insights into the trade-offs between training from scratch and using PVRs, informing the design of future visual RL algorithms.

\end{abstract}
\section{Introduction}

Reinforcement learning (RL) has made tremendous progress in recent years, with the development of algorithms that enable robots to learn complex behaviors from high-dimensional sensory inputs. In visual RL, where robots learn from raw image observations, a key challenge lies in extracting relevant features from the visual data. Despite significant advances in computer vision over the last decade, current state of the art models in visual RL still use simple convolutional neural networks (CNNs) trained from scratch for different tasks. This approach has been successfully employed in algorithms such as DRQ-V2 \cite{yarats2021mastering} and DRM \cite{xu2023drm}, which have solved some of the most challenging continuous control tasks in common RL benchmarks such as the humanoid run and dog tasks in the deepmind control suite \cite{xu2023drm}.

Notwithstanding these successes, training a CNN from scratch does not leverage the knowledge captured by pre-trained models on large datasets. Recently, there has been growing interest in using pre-trained image encoders as feature extractors for Imitation Learning (IL) and RL \cite{darcet2023vision}\cite{karamcheti2023language}. These pre-trained models have been trained on massive datasets and have learned to extract rich and generalizable features that can be useful for a wide range of downstream tasks \cite{yuan2022pre}\cite{darcet2023vision}\cite{oquab2023dinov2}. By leveraging pre-trained models, RL algorithms may learn more efficiently. 

Applying RL to robotics poses unique challenges however, particularly in terms of sample efficiency and control frequency. Specifically, data collection is often time-consuming and expensive as it requires physical interactions with the environment. Moreover, robots typically operate at high frequencies, requiring control signals to be generated at rates of 10-100 Hz or more. When training a real robot from scratch, this demands RL algorithms that can learn quickly and make decisions rapidly.

Despite the potential benefits of using pre-trained image encoders in RL, there are limited comparisons of the two methods in surrounding literature. In their investigation of an "artificial visual cortex" for embodied AI, Majumdar et al. \cite{majumdar2024we} compare pre-trained visual representations (PVRs) across multiple robotic manipulation and navigation tasks, but do not compare these models to a learning-from-scratch RL implementation. In their comparisons, they find that no single PVR is universally dominant across multiple domains. Parisi et al. \cite{parisi2022unsurprising} compare learning from scratch to using PVRs, however their study only considers learning from demonstrations using imitation learning. 

In this paper, we compare visual RL algorithms that train a CNN from scratch with counterparts that utilize frozen PVRs. We investigate the performance of these algorithms on a simulated robotic pushing task, and analyze the trade-offs between training a CNN from scratch and using pre-trained feature extractors.

\section{Related work}

\subsection{Visual Reinforcement Learning}

In domains such as robotics, choosing the right RL method to maximize learning efficiency is important for minimising the risk of robot degradation. In this context, off-policy methods such as SAC \cite{haarnoja2018soft} and DDPG \cite{lillicrap2015continuous} are generally more suitable than on-policy methods because off-policy methods can learn from experiences gathered without following the same policy that is being learned. This  allows them to leverage data collected from various sources such as demonstrations, simulations, or other robots. In contrast, on-policy methods such as policy gradient methods require data to be collected using the same policy that is being learned. This can be limiting in applications where data acquisition is expensive.

More recent advances in off-policy visual RL, such as the Data-Regularized Q (DRQ) \cite{kostrikov2020image} algorithm  and its successor, DRQ-V2 \cite{yarats2021mastering}, have improved the efficiency and effectiveness of off-policy learning in visual RL tasks by using random shift augmentations for regularisation. These methods have been shown to achieve state-of-the-art performance in a variety of simulated robotics tasks such as grasping and manipulation, from a relatively small amount of data. 

In this paper, we use the Dormant Ratio Minimization (DRM) \cite{xu2023drm} visual RL algorithm as the baseline. DRM was chosen because it is the most sample efficient RL algorithm for visual control and has exhibited very good performance on continuous control tasks \cite{xu2023drm}. DRM builds from the observation that there is a connection between the reduction of an agent's dormant ratio and the agent's skill acquisition in visual control tasks \cite{xu2023drm}. A dormant neuron is a neuron that has become nearly inactive, displaying a minimal activation level relative to other neurons in a layer. The dormant ratio is the percentage of these dormant neurons in the entire network \cite{xu2023drm}. DRM uses the dormant ratio to guide exploration, and periodically perturbs the agent's weights when the dormant ratio is high.

\subsection{Reinforcement Learning with Pre-trained Visual Representations}

Yuan et al. \cite{yuan2022pre} adapt the DRQ-v2 algorithm \cite{yarats2021mastering} to use an ImageNet pretrained ResNet18 as a PVR. The authors find that using features from early layers provides better performance for continuous control tasks when compared to use of the full network. Their approach outperforms DRQ-v2 on Deepmind control suite tasks with video backgrounds. 

Xiao et al. \cite{xiao2022masked} demonstrate that self-supervised pre-training from real world images is effective for learning motor control from pixels. They train a vision transformer (ViT) model by masked modelling of natural images. This visual encoder is then frozen and the RL policy is trained using the frozen encoder. Another finding of the paper is that "in the wild" images, such as those from YouTube or egocentric videos, lead to better visual representations for manipulation tasks when compared to ImageNet images. 

\subsection{Learning from Video Demonstrations}
In recent years, there has been a growing interest in using RL as a method of learning from video demonstrations (LfVD) \cite{schmeckpeper2020reinforcement}\cite{zakka2022xirl}. LfVD involves learning a policy from a set of video demonstrations, often collected from humans or other agents. This is motivated by the fact that learning policies from expert videos is a scalable way of learning a wide array of tasks without access to ground truth rewards or actions \cite{escontrela2024video}. By leveraging the knowledge and expertise encoded in these demonstrations, LfVD can significantly reduce the amount of data required to learn a task \cite{schmeckpeper2020reinforcement}\cite{zakka2022xirl}. However, LfVD also relies on the ability to extract relevant features from the video data. Sharing a common PVR for both LfVD and RL could enable more efficient learning.

Zakka et al. \cite{zakka2022xirl} use an ImageNet pretrained ResNet18 with a self-supervised Temporal Cycle Consistency loss to learn reward functions from video demonstrations. They demonstrate learning robot policies from videos of human demonstrations on the Metaworld Push environment. Whilst they use a PVR for learning the reward, they then use state based RL for learning from the videos. Finding a suitable PVR for visual RL could allow both the visual state observation and reward to be given by the PVR when using reward functions learned from videos.

\section{Methodology}

\subsection{Benchmarking}

We use the Push-v2 task from the Metaworld benchmarking suite for our evaluation. Metaworld consists of 50 robotic manipulation tasks such as opening and closing doors and drawers and pressing buttons \cite{yu2020meta}. The chosen task consists of pushing a small red puck towards a green sphere, with the initial puck and sphere positions randomised at the beginning of each episode. An episode is considered a success when the puck is within 5cm of the target at the final timestep. The reward function rewards minimizing the distance from the gripper to the puck, from the puck to the target site, and gives a bonus when the object is grasped by the robot. A maximum reward of 20 is given when the object is within 5cm of the target site. 

From the Metaworld suite, this task was chosen as it involves manipulating a small object which could be contained entirely within one patch of a ViT. It was also selected due to it being one of the hardest tasks to solve in the Metaworld suite according to the original paper \cite{yu2020meta}. One explanation for the task's difficulty could be that the puck can fall on its side and roll away from the target position if not carefully placed, resulting in a failed episode even if the object temporarily reaches the target zone.

For further evaluation we use the Drawer-Open-v2 task. The task consists of opening a drawer, with the drawer position randomised at the beginning of each episode. The reward function rewards minimizing the distance of the gripper to the handle, and the distance of the handle to a target position. An episode is considered a success when the handle is less than 3cm from a target position and the gripper is less than 3cm from the handle. This task was chosen due to it involving a larger and more natural looking object.

\begin{figure}
    \centering
    \subfloat[Push-v2]{\includegraphics[width=0.4\textwidth]{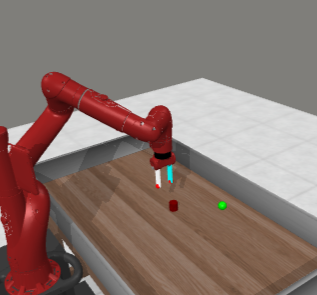}}
    \hfill
    \subfloat[Drawer-Open-v2]{\includegraphics[width=0.37\textwidth]{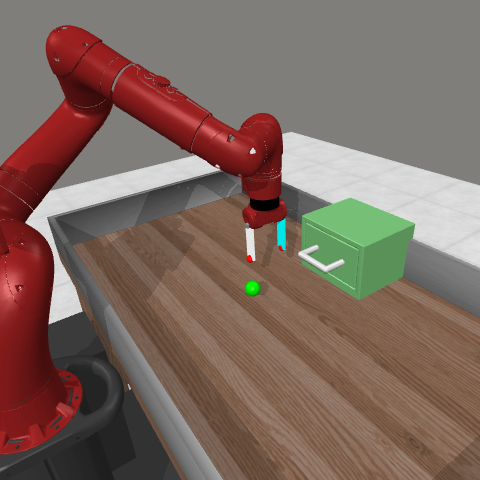}}
    \caption{Metaworld Tasks}
    \label{fig:enter-label}
\end{figure}

All experiments use an image resolution of 112x112 and maintain a consistent third-person camera position unless otherwise noted. The default DRM hyperparameters \cite{xu2023drm} are used for all experiments. 

\subsection{Pretrained Visual Representations}

We compare the baseline DRM algorithm to three PVRs: ResNet18, DINOv2, and Visual Cortex (VC). The PVR's weights are frozen throughout the RL training. 

\subsubsection{ResNet18}
An ImageNet pretrained ResNet18 was chosen due it being a widely used and well established CNN that has demonstrated its efficacy in a variety of computer vision tasks such as classification, object detection and segmentation. Other reasons for its use are its relatively small size and its high performance in visual RL, shown in "Pre-Trained Image Encoder for Generalizable Visual Reinforcement Learning" (PIEG) \cite{yuan2022pre}. 

We adopt a similar approach to PIEG, where the images are fed through the first two layers of a ResNet18 CNN. To mitigate the high dimensionality of the output, we flatten the CNN output and pass it through a trainable fully connected layer with a reduced output size. Consistent with DRM's CNN encoder, we omit pooling layers after the CNN output to preserve spatial information. However, this design choice comes with a limitation: higher resolutions would result in an explosion of parameters in the fully connected layer, making it computationally prohibitive. The output of this layer is then fed into the actor and critic networks of DrM.

\subsubsection{DINOv2}
DINOv2 is a self-supervised vision backbone based on a ViT model. It is trained using a student teacher approach at both a whole image level and at a patch level. A 1 billion parameter model is trained on a curated dataset of 142 million images. This large model is then distilled down into smaller models. The backbone gives state of the art results for self-supervised networks in downstream tasks such as image classification, semantic segmentation and depth estimation. In this paper we utilise the 86 million parameter ViT-B model. DINOv2 was chosen as a PVR for comparison as it is one of the best performing self-supervised ViT based models for downstream tasks noted above.

Darcet et al. \cite{darcet2023vision} find that artefacts in transformer feature maps correspond to high norm patch tokens appearing in low information background areas of images. The authors hypothesize that large, sufficiently trained models recognize these redundant tokens and use them as places to store, process and retrieve global information. By adding additional tokens (register tokens) to the transformer input to store and process global information, these high norm patch patch tokens disappear. 

To use DINOv2 as a PVR, we pass the image through the model and feed the CLS token directly into the actor and critic networks. We also compare this to concatenating the register tokens to the CLS token and passing these concatenated tokens into the actor and critic networks.

\subsubsection{Visual Cortex}
Visual cortex (VC) \cite{majumdar2024we} is a ViT based PVR model trained using Masked Auto-Encoding (MAE) on egocentric data, with the aim of being used for embodied intelligence. VC was chosen for this comparison because it was designed for use in embodied AI. It was also found to be the best performing PVR on average in comparisons performed in \cite{majumdar2024we}. As with DINOv2, we feed the CLS token directly into the actor and critic networks.

\subsection{Visual RL Algorithm}
As we focus on the performance of the vision systems, we do not include robot proprioception information in the state space. To compare between the baseline and the pretrained backbones, the CNN based encoder is replaced with the pretrained backbone.

DRM stores observations as images in the replay buffer. At each update step a batch of these observations is augmented and passed through the image encoder, actor and critic networks. When using the ViT based PVRs, the image observations are passed through the frozen vision backbone at each environment step. The CLS token, which is a learnable vector that serves as a representation of the entire image \cite{dosovitskiy2020image}, is given by the backbone and  is stored in the replay buffer. This reduces the memory footprint of the replay buffer by over 90\%, since the CLS token is a 768 dimensional float 32 vector, compared to the 3x112x112 8 bit integer image. Passing one image through the backbone at each time step also avoids the need to pass mini batches of 256 through the backbone at each update step. 

Another advantage of storing the CLS tokens in the replay buffer is that the size of the replay buffer does not increase with higher resolutions, since only the CLS token is stored. The original DRM \cite{xu2023drm} paper and PIEG \cite{yuan2022pre} use image observations of size 84x84 with a replay buffer size of 1M. Uncompressed, this results in a replay buffer size of 21.17 GB. At a resolution of 112x112, this increases to 37.63GB, and at 224x224 (the resolution commonly used by ViT models), this increases to 150.53GB. These large replay buffers pose challenges to the training of CNNs from scratch for visual RL. 

Since no pooling layers are used in DRM's image encoder, the number of parameters in the actor and critic network increases with the input resolution. This limits the maximum resolution that can be used while still being able to run at a frequency suitable for robotic control. Table \ref{tab:trainable-params} shows how the number of parameters differs at different resolutions.

\begin{table}
\centering
\caption{Trainable Parameters}
\label{tab:trainable-params}
\begin{tabular}{lrr}
\hline
\textbf{Model} & \textbf{112x112} & \textbf{224x224} \\
\hline
DRM & 15978281 & 57373481 \\
PIEG & 107340233 & 415621577 \\
DINOv2 & 4768713 & 4768713 \\
DINOV2 REG & 6151113 & 6151113 \\
VC & 4768713 & 4768713 \\
\hline
\end{tabular}
\end{table}

\subsection{Performance Metrics}
We present graphs of the episode reward and success rate of the evaluation episodes. The episode reward is the total reward obtained by the agent for an entire episode, and the success rate is the proportion of evaluation episodes where the agent successfully completed the task. All results are plotted as the mean performance over 5 seeds with a standard deviation shading of $\pm$ 0.5.

\section{Experiments and Results}

\subsection{Data Augmentation}

DRM applies a padding and random shift augmentation to all images before they are passed through the encoder to regularize the input data \cite{yarats2021mastering}\cite{xu2023drm}. We first test whether this augmentation is beneficial when using PVRs which have been trained on large datasets with augmentations \cite{oquab2023dinov2}. We compare the results of the DINOv2 CLS token on the Push-v2 task with and without the random shift augmentation. 

\begin{figure}
    \centering
    \subfloat[Rewards]{\includegraphics[width=0.48\textwidth]{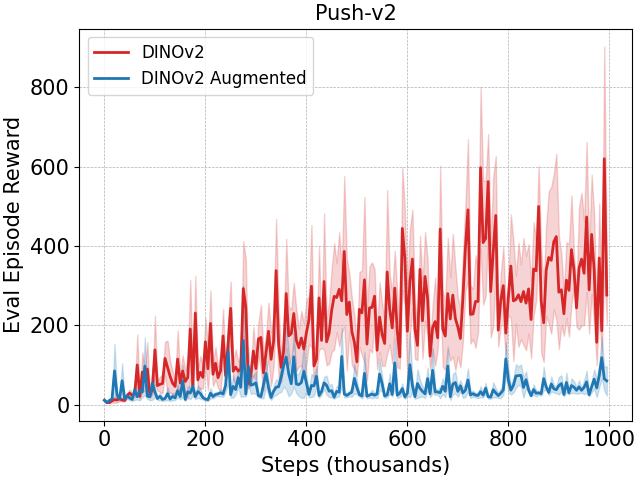}}
    \hfill
    \subfloat[Success Rate]{\includegraphics[width=0.48\textwidth]{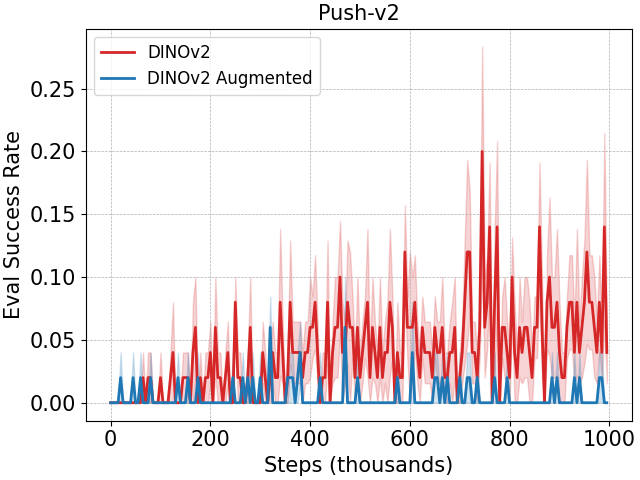}}
    \caption{Comparison of using plain images to using random shift augmentations.}
    \label{fig:aug}
\end{figure}

As shown in figure \ref{fig:aug}, the DINOv2 PVR performed better when the random shift augmentations used in the baseline DRM algorithm were not applied. Therefore, we do not apply these augmentations to images passed into the ViT based PVRs that are included in this comparative study.

\subsection{PVR Performance}

\begin{figure}[ht]
    \centering
    \subfloat[Rewards]{\includegraphics[width=0.32\textwidth]{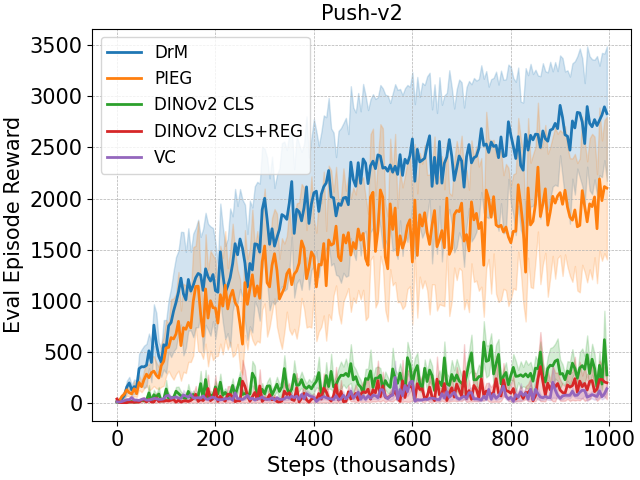}}
    \hfill
    \subfloat[Success Rate]{\includegraphics[width=0.32\textwidth]{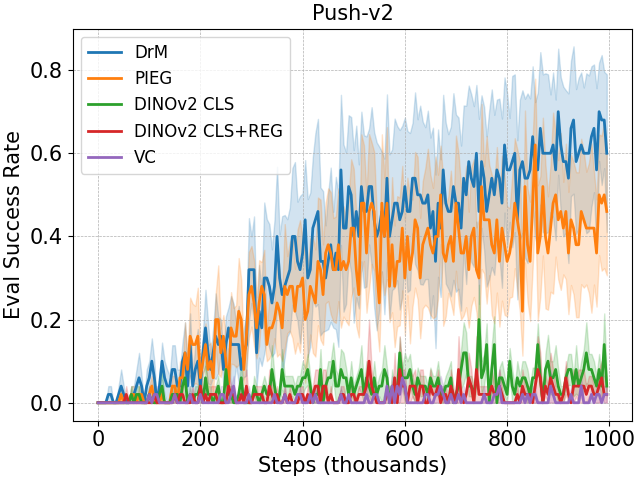}}
    \hfill
    \subfloat[Dormant Ratio]{\includegraphics[width=0.34\textwidth]{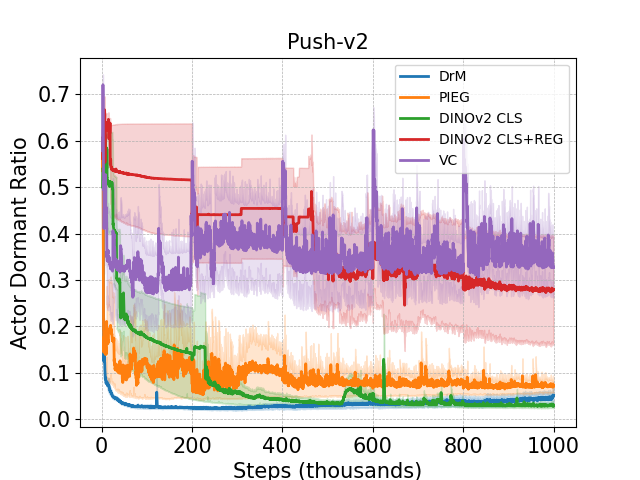}}
    \caption{Mean performance on Push-v2 Metaworld task.}
    \label{fig:performance}
\end{figure}

\begin{figure}[ht]
    \centering
    \subfloat[Rewards]{\includegraphics[width=0.32\textwidth]{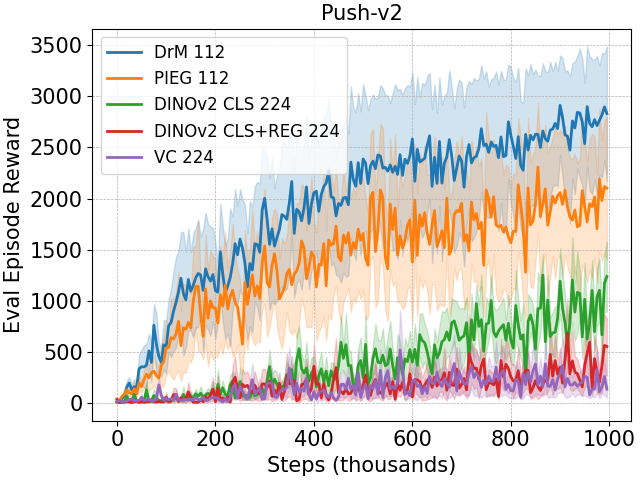}}
    \hfill
    \subfloat[Success Rate]{\includegraphics[width=0.32\textwidth]{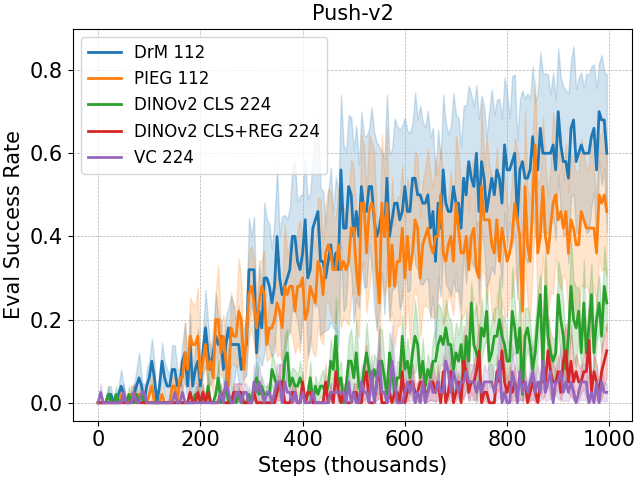}}
    \hfill
    \subfloat[Dormant Ratio]{\includegraphics[width=0.32\textwidth]{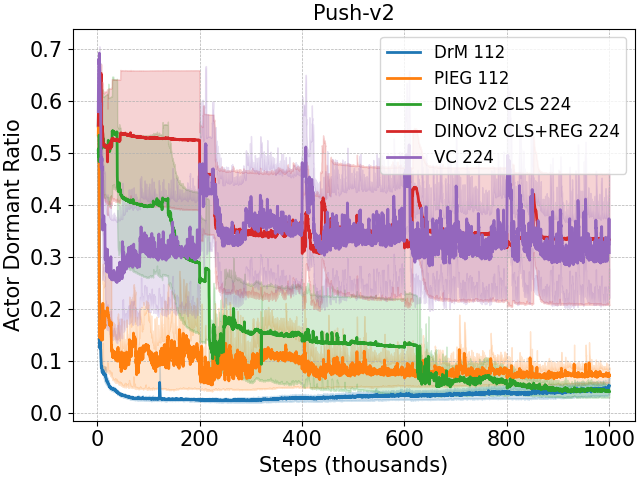}}
    \caption{Mean performance on Push-v2 Metaworld task with increased resolutions for ViT PVRs.}
    \label{fig:performance_224}
\end{figure}

\begin{figure}[ht]
    \centering
    \subfloat[Rewards]{\includegraphics[width=0.32\textwidth]{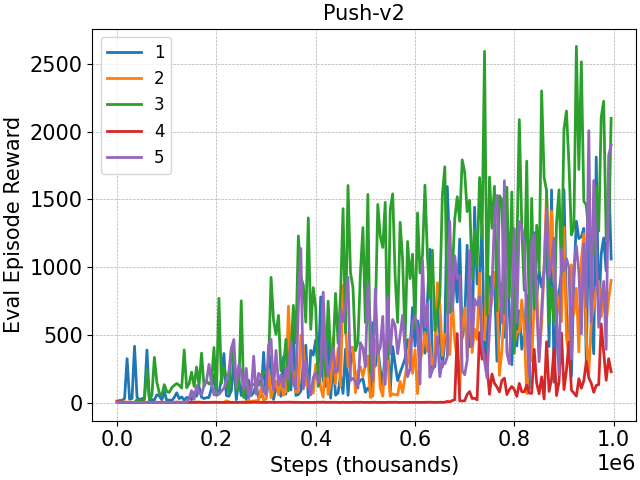}}
    \hfill
    \subfloat[Success Rate]{\includegraphics[width=0.32\textwidth]{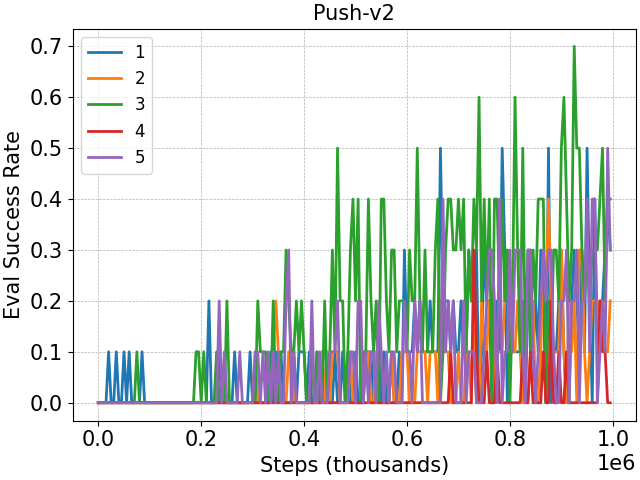}}
    \hfill
    \subfloat[Actor Dormant Ratio]{\includegraphics[width=0.32\textwidth]{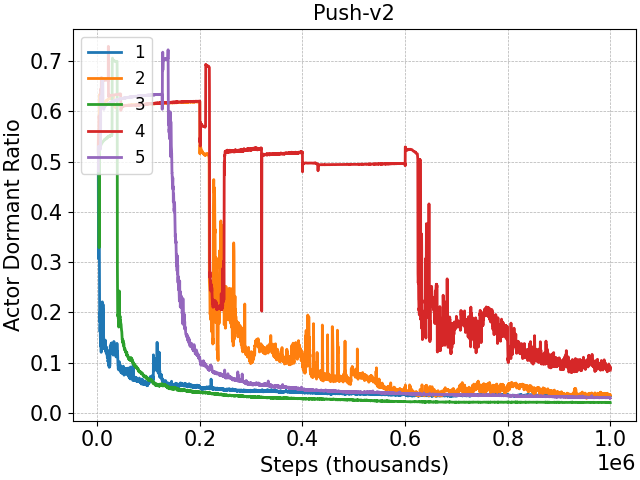}}
    \caption{Per Seed Performance of DINOv2 CLS token at 224x224 resolution}
    \label{fig:seeds}
\end{figure}

Figure \ref{fig:performance} presents the performance of the different models on the Push-v2 task. The baseline DRM algorithm achieves the best performance, closely followed by the PIEG-based ResNet model. In contrast, the ViT-based models exhibit significantly poorer performance, highlighting a substantial gap between CNN-based and ViT-based models.

Figure \ref{fig:performance} also reveals a strong correlation between the dormant ratio and model performance. The top-performing models, including the DRM and PIEG-based ResNet, demonstrate a drastic reduction in the dormant ratio throughout training. In contrast, the poorly performing models, including the ViT-based models, maintain a high dormant ratio throughout training.

To explore the potential benefits of using ViT-based models at higher resolutions, we additionally tested these models at a resolution of 224x224. Whilst the CNN-based models are unable to operate at this resolution due to the computational constraints of the large replay buffer size, the ViT-based models can process images at this resolution with only a marginal decrease in control frequency.

As shown in figure \ref{fig:performance_224}, the performance of the ViT models improves slightly at the higher resolution, with the DINOv2 CLS token achieving the best performance. However, even at this higher resolution, the ViT models are still significantly outperformed by the CNN-based models at the lower resolution.

Figure \ref{fig:seeds} illustrates the high variance in performance across different seeds for the DINOv2 CLS model at the 224x224 resolution. Notably, the worst performing seeds are those that fail to reduce their dormant ratio throughout training. This trend is consistent across the ViT-based models, with the VC model exhibiting only one convergent seed out of five, and the DINOv2 with register tokens showing three convergent seeds out of five.

\begin{figure}[ht]
    \centering
    \subfloat[Rewards]{\includegraphics[width=0.44\textwidth]{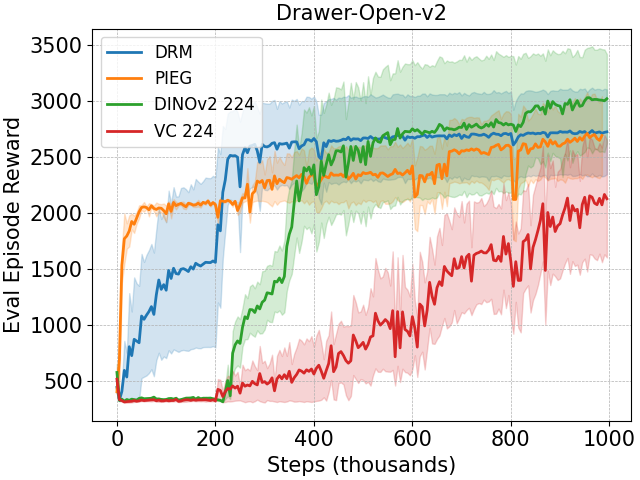}}
    \hfill
    \subfloat[Success Rate]{\includegraphics[width=0.44\textwidth]{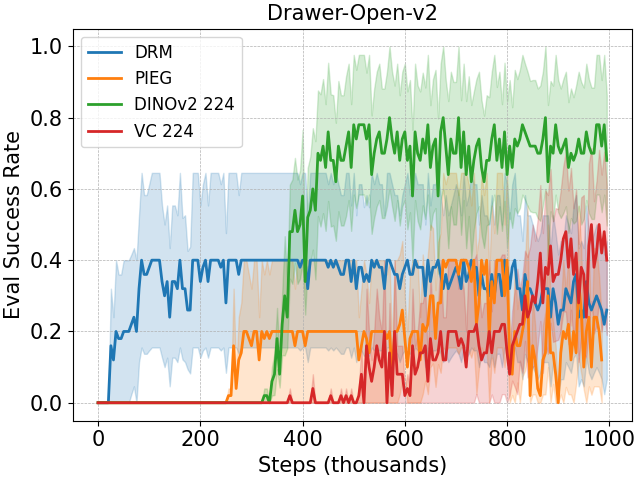}}
    \caption{Mean performance on  Drawer-Open-v2 Metaworld task with increased resolutions for ViT PVRs.}
    \label{fig:drawer_graphs}
\end{figure}

To investigate whether these results are task specific, we compare the ViT models at 224x224 to the CNN models at 112x112 on a drawer opening task. The results of this experiment are shown in figure \ref{fig:drawer_graphs}. We find that the DINOv2 PVR outperforms both DRM and PIEG. This could be explained by the drawer providing more visual features due to its size and shape when compared to the rendered cylinder and sphere in the pushing task. The large difference between the rewards and success rate are due to some seeds ending stuck in a local minima of grasping the handle without opening the door, indicating the need for further exploration.

\begin{figure}[ht]
    \centering    \subfloat{\includegraphics[width=0.19\textwidth]{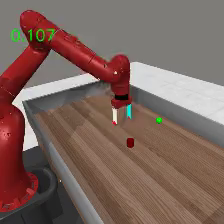}}
    \hfill
    \subfloat{\includegraphics[width=0.19\textwidth]{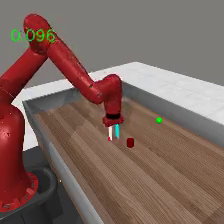}}
    \hfill
    \subfloat{\includegraphics[width=0.19\textwidth]{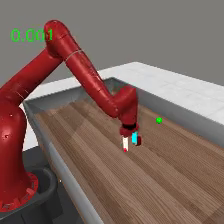}}
    \hfill
    \subfloat{\includegraphics[width=0.19\textwidth]{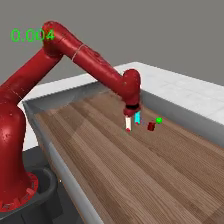}}
    \hfill
    \subfloat{\includegraphics[width=0.19\textwidth]{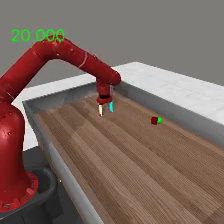}}
    \vfill
    \subfloat{\includegraphics[width=0.19\textwidth]{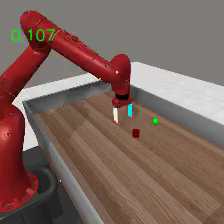}}
    \hfill
    \subfloat{\includegraphics[width=0.19\textwidth]{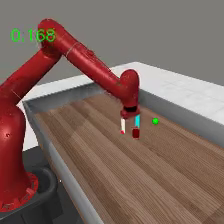}}
    \hfill
    \subfloat{\includegraphics[width=0.19\textwidth]{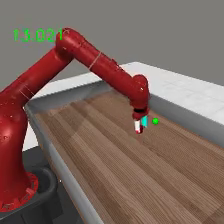}}
    \hfill
    \subfloat{\includegraphics[width=0.19\textwidth]{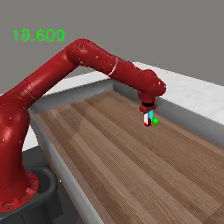}}
    \hfill
    \subfloat{\includegraphics[width=0.19\textwidth]{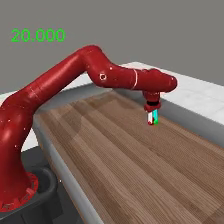}}
    \caption{Examples of different learned approaches. Top: robot hits puck towards target. Bottom: Robot grabs object and moves it to target.}
    \label{fig:push_tactics}
\end{figure}

The videos of the evaluation episodes show that the robot sometimes learned different strategies to complete the task. Some agents would try to hit the puck towards the target with the outside of the gripper, whilst others would use the gripper to grasp the puck and place it near the target. These strategies are shown in figure \ref{fig:push_tactics}.

\section{Discussion}

Our comparison reveals that the choice of training from scratch or using a PVR for maximising performance in visual RL is task dependant. With the DRM algorithm performing best on the pushing task and the DINOv2 PVR performing best on the drawer opening task, we hypothesize that PVRs perform better when the objects being manipulated are larger and more distinguishable. We also identify several advantages of using PVRs, including a significant reduction in replay buffer size and operating at a higher control frequency. 

We find that on the push-v2 task the PIEG PVR outperforms other PVRs, and the DINOv2 CLS token is the top-performing ViT-based PVR. Interestingly, our results indicate that the DINOv2 ViT outperforms the Visual Cortex (VC) model, which was specifically designed for embodied AI. Whilst Darcet et al. \cite{darcet2023vision} demonstrate the effectiveness of register tokens in training ViTs to avoid storing image-level information in patch tokens, we find that including these register tokens in the state space does not improve visual RL performance when compared to sole use of the CLS token.

Furthermore, our analysis highlights the importance of the dormant ratio in visual RL. We found that a lower dormant ratio is a strong indicator of successful learning, and the best-performing models demonstrate a drastic reduction in the dormant ratio throughout training. In contrast, poorly performing models maintained a high dormant ratio, suggesting that the robot was unable to effectively explore and learn from the environment. The PVR based approaches resulted in a higher number of failed seeds where the dormant ratio did not decrease throughout training. This finding has significant implications for incorporating PVRs into visual RL algorithms, as it suggests that incorporating alternative mechanisms to reduce the dormant ratio may be essential for achieving successful learning outcomes.

Using PVRs also reduces the number of trainable parameters, which can lead to faster training times and lower computational costs. This is crucial for tasks such as robotic manipulation and navigation in the real world, where data collection is costly and time-consuming. This trade-off between performance and complexity is important in visual RL, where large models can be prohibitive for real time control. While PVRs may not yet always match the performance of training from scratch, they offer a promising approach for balancing performance and complexity.

\section{Conclusion}

Our study provides a comprehensive analysis of the effectiveness of pre-trained PVRs in visual RL. Our results demonstrate that the optimal performance of PVRs or from scratch learning is task-dependent and highlight the importance of evaluating visual RL algorithms across multiple tasks and domains. We identify several advantages of using PVRs, including reduced replay buffer size and computational costs. We also find that the general purpose DINOv2 PVR outperforms the Visual Cortex PVR that was specifically designed for use in embodied AI. Our findings suggest that mechanisms to reduce the dormant ratio may be essential to achieve successful learning outcomes when incorporating PVRs into visual RL algorithms.

Future work will explore alternative RL algorithms, regularization approaches for PVRs, analysis across further tasks, and fine-tuning PVRs during training to provide a more comprehensive understanding of the strengths and limitations of PVRs in visual RL.

\begin{credits}
\subsubsection{\ackname} This work was supported by the Engineering and Physical Sciences Research Council [EP/S023917/1] and Dogtooth Robotics.
\end{credits}
%
%
%
\bibliographystyle{splncs04}
\bibliography{references}
\end{document}